\documentclass[letterpaper, 10pt, conference]{ieeeconf}
\IEEEoverridecommandlockouts % to override locked commands
\usepackage{cite}
\usepackage{color}
\usepackage{graphicx}
\usepackage{amsmath}
\usepackage{amssymb}
\usepackage{listings}
\usepackage{algorithmic}
\usepackage[ruled]{algorithm2e}
\usepackage[nonumberlist,acronym]{glossaries}
\usepackage{array}
\usepackage{placeins}
\usepackage{stfloats}
\usepackage{url}
\usepackage{multirow}
\usepackage{multicol}
\usepackage[table,xcdraw]{xcolor}
\usepackage{soul}
\usepackage{float}

\usepackage{mathtools}

\newcolumntype{P}[1]{>{\centering\arraybackslash}p{#1}}
\SetKwRepeat{Do}{do}{while}%

\usepackage{ctable}

\graphicspath{{./figures/}}

\newacronym{ROS}{ROS}{Robot Operating System}
\newacronym{IMU}{IMU}{Inertial Measurement Unit}
\newacronym{GPS}{GPS}{Global Positioning System}
\newacronym{CLAHE}{CLAHE}{Contrast Limited Adaptive Histogram Equalization}
\newacronym{SSC}{SSC}{Suppression via Square Covering}
\newacronym{SLAM}{SLAM}{Simultaneous Localization and Mapping}
\newacronym{RANSAC}{RANSAC}{Random Sampling Consensus}
\newacronym{SURF}{SURF}{Speeded-up Robust Features}
\newacronym{AHE}{AHE}{Adaptive Histogram Equalization}
\newacronym{SDC}{SDC}{Supression via Disk Covering}
\newacronym{VO}{VO}{Visual Odometry}
\newacronym{SFM}{SFM}{Structure from Motion}
\newacronym{SIFT}{SIFT}{Scale Invariant Features Transform}
\newacronym{ORB}{ORB}{Oriented FAST and Rotated BRIEF}
\newacronym{AOR}{AOR}{Angle based Outliers Rejection}

\begin{document}

\title{\LARGE \bf Drift Reduction for Monocular Visual Odometry of Intelligent Vehicles using Feedforward Neural Networks} % A working title (to be further modified)

%A Machine Learning Approach for Orientation Drift Reduction in Monocular Visual Odometry for Intelligent Vehicles

\author{Hassan Wagih$^{1*}$, Mostafa Osman$^{2*}$, Mohammed I. Awad$^{1}$, and Sherif Hammad$^{1}$%, Mohamed W. Mehrez$^{2}$, Soo Jeon$^{2}$, and William Melek$^{2}$ % <-this % stops a space
	\thanks{$^{1}$ Mechatronics Department, Ain Shams University, Cairo 11566, Egypt
		{\tt\small \{Hassan.Wageh, Mohammed.awad\}@eng.asu.edu.eg}\tt\small\ ,Sherif.hammad@garraio.com}%
	\thanks{$^{2}$ Mechanical, Materials, and Maritime Engineering, Delft University of Technology, Delft, The Netherlands
		{\tt\small M.E.A.Osman@tudelft.nl}}%
	\thanks{$^{*}$ The first two authors contributed equally to this work.}%
}

\maketitle

\begin{abstract}
In this paper, an approach for reducing the drift in monocular visual odometry algorithms is proposed based on a feedforward neural network. A visual odometry algorithm computes the incremental motion of the vehicle between the successive camera frames, then integrates these increments to determine the pose of the vehicle.
The proposed neural network reduces the errors in the pose estimation of the vehicle which results from the inaccuracies in features detection and matching, camera intrinsic parameters, and so on. These inaccuracies are propagated to the motion estimation of the vehicle causing larger amounts of estimation errors.
The drift reducing neural network identifies such errors based on the motion of features in the successive camera frames leading to more accurate incremental motion estimates.
The proposed drift reducing neural network is trained and validated using the KITTI dataset and the results show the efficacy of the proposed approach in reducing the errors in the incremental orientation estimation, thus reducing the overall error in the pose estimation.
\end{abstract}

%%%%%%%%%%%%%%%%%%%%%%%%%%%%%%%%%%%%%%%%%%%%%%%%%%%%%%%%%%%%%%%%%%%%%%%%%

%\begin{IEEEkeywords}
%Visual Odometry, Machine Learning, Neural Networks, Localization, Self-driving Vehicles
%\end{IEEEkeywords}

\section{Introduction}
\label{sec:introduction}
The interest in developing self-driving vehicles capable of navigating their environments without human intervention has significantly grown during the last decade. One of the main modules of a self-driving vehicle is the localization module~\cite{siegwart2011introduction}. A self-driving vehicle has to be able to estimate its position and orientation accurately in order to navigate its environment successfully. To achieve accurate localization, several exteroceptive sensors are usually used such as Global Positioning Systems (GPS)~\cite{drawil2012gps}, Cameras~\cite{leutenegger2015keyframe}, Light Detection And Ranging (LiDAR)~\cite{zhang2014loam}, and so on.

Cameras can be utilized for localization using  Visual Odometry (VO) algorithms~\cite{scaramuzza2011visual, fraundorfer2012visual}. A VO algorithm estimates the incremental motion between successive frames captured by the camera. The position and orientation of the vehicle are then estimated by integrating these motion increments at each timestep. VO algorithms can be classified based on the type of camera into monocular~\cite{sabry2019ground}, stereo~\cite{fan2017stereo}, and RGB-D~\cite{sturm2012benchmark} visual odometries~\cite{aqel2016review}. It can also be classified based on the motion estimation approach as feature-based and direct VO. Feature-based VO relies on detecting and matching visual features in the successive frames. Such features are then used to solve the motion estimation problem.

A monocular visual odometry algorithm uses the captured frames from only one camera to estimate the motion of the vehicle. Such technique uses Structure from Motion (SFM) to estimate the incremental motion~\cite{tomono20053, hartley2003multiple}. However, since the algorithm uses only one camera, the incremental motion of the vehicle can only be estimated up to an unobservable scale which can be determined using another sensor such as a wheel encoder or an Inertial Measurement Unit (IMU). 

Since VO algorithms rely on integration in estimating the position and orientation, %{the accumulation of error every time-step increases with time due to the individual errors in each incremental motion}.
individual errors in each incremental motion lead to the accumulation of error in every time step.
Although the drift error in VO algorithms cannot be eliminated, several works have tried to reduce such drift. In~\cite{desai2016visual}, a  new descriptor called the Syntheitc BAsis (SYBA) descriptor is used to reduce the falsely matched features between the successive frames. In order to achieve this, a sliding window approach is developed where the features detected in a given frame is not only matched to the prior frame but to a window of previously captured frames. In~\cite{peretroukhin2017reducing}, the drift in the estimation of a VO algorithm is reduced using a Bayesian Convolutional Neural Network that infers the direction of the sun. The sun direction then provides a global estimate of the orientation estimate which consequently reduces the drift in the visual odometry output. 

Another approach to reduce the drift in the estimation of VO is through using sensor fusion techniques where the estimate from the VO algorithm is fused with other estimates from other sensors to reduce the error. Sensor fusion relies on probabilistic estimation approaches such as Extended Kalman Filter~\cite{thrun2002probabilistic}, Moving Horizon Estimation~\cite{osman2021generic} and pose graph optimization~\cite{tao2021multi}, and so on. In such approaches, the uncertainty in the position and orientation estimates need to be quantified using the covariance matrix of the measurements in order to enhance the accuracy of the estimation. In~\cite{osman2019novel, osman2018online}, a covariance estimation approach was developed utilizing a sensor not suffering from drift to estimate the covariance of the drift suffering odometry.

In this paper, a machine learning approach is proposed for reducing the drift in the incremental orientation estimates of a monocular VO algorithm which in turn results in the overall enhancement of the integrated position and orientation accuracy. By developing a neural network that can correlate the errors in the motion increments estimated by the VO algorithm with the motion of the features in the successive frames, the error in such increments can be reduced significantly. By using an accurate Realtime Kinematic GPS (RTK-GPS) during the training phase, the neural network can be trained to estimate the error in the visual odometry increments which can be then used to reduce the motion error in realtime during the operation of the vehicle. The proposed approach is validated using sequences from KITTI dataset with its ground-truth data~\cite{geiger2013vision}. 

The remainder of the paper is organized as follows, Section~\ref{sec:drift_modeling} contains the mathematical formulation of the drift error in the visual odometry followed by our proposed approach for reducing the drift in Section~\ref{sec:data_driven}. In Section~\ref{sec:experimental_work} the experimental work executed to evaluate the proposed method is explained and the results are shown in Section~\ref{sec:results_discussion}. Finally, the conclusion and future work are stated in Section~\ref{sec:conclusion}. 

%%%%%%%%%%%%%%%%%%%%%%%%%%%%%%%%%%%%%%%%%%%%%%%%%%%%%%%%%%%%%%%%%%%%%%%%%

\section{Error Modeling in Visual Odometry}
\label{sec:drift_modeling}

\begin{figure*}
	\centering
	\includegraphics[width=0.9\linewidth]{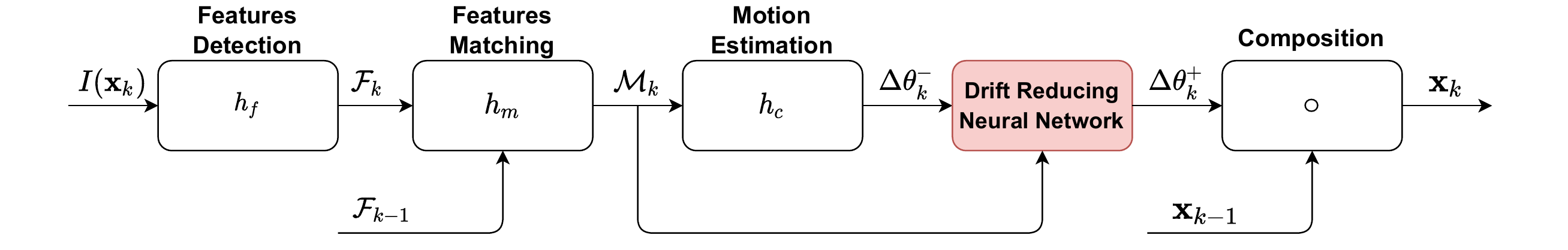}
	\caption{The overall feature-based visual odometry pipeline with the proposed drift reducing neural network. The image $I$ at the position and orientation $\mathbf{x}_{k}$ is passed to a features detection algorithm $h_{f}$ followed by a matching algorithm $h_{m}$ which matches the detected features to the features from the previous frame. The matched features are then used to estimate the motion increment using a motion estimation algorithm $h_{c}$. The statistical moments of the features motion in the camera frames are then passed to the NN along with the motion increment from the VO algorithm. Finally, the NN outputs a more accurate motion increment which is then used to calculate the current position and orientation of the vehicle through composition} 
	\label{fig:overall_pipeline}
\end{figure*}

As mentioned earlier, a visual odometry estimate the position and orientation of a vehicle by integrating the incremental motion. For feature-based VO algorithm, the first step is detecting the features in the frame, this can be modeled as 
\begin{equation}
	\label{equ:feature_model}
	\mathcal{F}_{k} = \underbrace{h_{\text{f}}(\mathbf{x}_{k})}_{\hat{\mathcal{F}}_{k}} + \mathbf{w}_{k},
\end{equation} 
where $\mathbf{x}_k = [\mathbf{p}^{\top}_k, \boldsymbol{\theta}^{\top}_k] \in \mathbb{R}^{3} \times \mathbb{M}$ is the state vector of the vehicle in the $k$-th time-step, which is composed of the position $\mathbf{p} \in \mathbb{R}^{3}$ and the orientation $\boldsymbol{\mathbf{\theta}} \in \mathbb{M}$ of the vehicle using an orientation representation (e.g. quaternions or Euler angles). $\mathcal{F}_{k} := \{f_{k}^{i}\}_{i=1}^{N_f} \in \mathbb{I}$ is the detected features set and $f := [u, v]^{\top}$ denotes a detected feature in the image. $h_{\text{f}}: \mathbb{R}^{3} \times \mathbb{M} \rightarrow \mathbb{I}$ is the model of the feature detection, and finally $\mathbf{w}_{k}$ is a noise term that models the errors in the feature detection process due to the noise affecting the camera. 

\begin{figure}[t]
	\centering
	\includegraphics[trim={1.5cm 4cm 4.8cm, 1.5cm}, clip, width=0.95\linewidth]{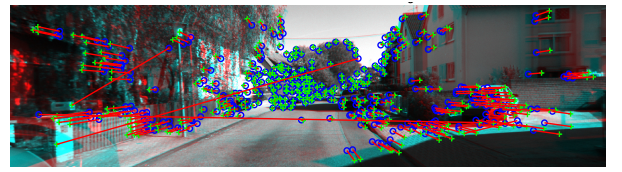}  
	\caption{An example of errors in mismatching of features between the 2180-th frame and the 2181-st frame in KITTI data set  sequence 5. The two frames are shown in the figure using red and blue shades. The features in the first frame is represented by the green (+) marks and blue circles in the second frame. Finally, the matched features are joined by red lines.}
	\label{fig:mistmatch}
\end{figure}

After the detection of features in the $k$-th frame, they are matched with the features detected in the $(k-1)$-th frame. Using \eqref{equ:feature_model} and applying Taylor expansion, this step can be modeled as
\begin{align}
	\label{equ:matching}
	\begin{split}
	\mathcal{M}_{k} &= h_{\text{m}}(\mathcal{F}_{k-1}, \mathcal{F}_{k}) + \varepsilon_{k}^{\text{m}} \\
	&\approx \underbrace{h_{\text{m}}(\hat{\mathcal{F}}_{k-1}, \hat{\mathcal{F}}_{k})}_{\hat{\mathcal{M}}_{k}}  +  
	\underbrace{\frac{\partial h_{\text{m}}}{\partial \hat{\mathcal{F}}_{k-1}} \mathbf{w}_{k-1} + \frac{\partial h_{\text{m}}}{\partial \hat{\mathcal{F}}_{k}} \mathbf{w}_{k}}_{\mathbf{w}^{\text{m}}_{k}} + \varepsilon_{k}^{\text{m}}, \\ 
%	&= \underbrace{h_{\text{m}}(\hat{\mathcal{F}}_{k-1}, \hat{\mathcal{F}}_{k})}_{\hat{\mathcal{M}}_{k}}  + \mathbf{w}^{\text{m}}_{k} + \varepsilon_{k}^{\text{m}},
	\end{split}
\end{align} 
where $\mathcal{M}_{k} := \{(f_{k-1}^{i}, f_{k}^{i})\}_{i=1}^{N_f} \in \mathbb{I}^{2}$ is the set of matched features (here we ignore all unmatched features), $h_{\text{m}}: \mathbb{I} \times \mathbb{I} \rightarrow \mathbb{I}^{2}$ is a mapping that represents the features matching algorithm, $\mathbf{w}^{\text{m}}_{k}$ is the effect of the white noise $\mathbf{w}_{k}$ on the matching, and $\varepsilon^{\text{m}}_{k} := \varepsilon^{\text{m}}(\mathcal{F}_{k-1}, \mathcal{F}_{k})$ is another noise term that is correlated to errors in the detection and matching of features in the successive frames. Such errors can result from false feature matching (wrong association), errors in the calibration of the intrinsic parameters of the camera, ... etc~\cite{yang2018challenges, calibration2008}. An example of matching errors in the VO pipeline is shown in Fig.~\ref{fig:mistmatch}. 
%where only unique features are matched, as the matching function performs forward-backward match to ensure that the matched features are unique (for more information, see Section~\ref{sec:experimental_work}).

Finally, depending on the type of camera being used, a motion estimation algorithm can be used to compute the motion increment from the matched feature pairs $\mathcal{M}_{k}$. This can be modeled similar to~\eqref{equ:matching} as 
\begin{align}
	\begin{split}
	\Delta \mathbf{x}_{k} &= h_{\text{c}}(\hat{\mathcal{M}}_{k}) + \frac{\partial h_{\text{c}}}{\partial \hat{\mathcal{M}}_{k}} (\mathbf{w}^{\text{m}}_{k} + \varepsilon_{k}^{\text{m}}) \\ 
	&= h_{\text{c}}(\hat{\mathcal{M}}_{k}) + \mathbf{w}_{k}^{\text{c}} + \varepsilon_{k}^{\text{c}},
	\end{split}
\end{align}
where $h_{\text{c}}: \mathbb{I}^{2} \rightarrow \mathbb{R}^{3} \times \mathbb{M}$ is the motion estimation function, $\mathbf{w}_{k}^{\text{c}}$ is the colored noise resulting from the origin white noise in~\eqref{equ:feature_model}, and $\varepsilon_{k}^{\text{c}} := \varepsilon^{\text{c}}(\mathcal{F}_{k-1}, \mathcal{F}_{k})$ is the effect of the noise term $\varepsilon_{k}^{\text{m}}$ on the motion estimation module. 

We aim at reducing the effect of the noise term $\varepsilon_{k}^{\text{c}}$ in order to increase the accuracy of the motion increments and ultimately reduce the drift in the VO algorithm. Using a Bayesian approach, this problem can be stated as the problem of estimating the distribution
\begin{equation}
	\label{equ:bayes}
	p(\Delta \mathbf{x}_{k} \mid \varepsilon_{k}^{\text{c}}, \mathcal{M}_{k}) = \eta \  p(\varepsilon_{k}^{\text{c}} \mid \Delta \mathbf{x}_{k}, \mathcal{M}_{k}) p(\Delta \mathbf{x}_{k} \mid \mathcal{M}_{k}). 
\end{equation}

In~\eqref{equ:bayes}, the prior distribution $p(\Delta \mathbf{x}_{k} \mid \mathcal{M}_{k})$ is the visual odometry algorithm. The distribution $p(\varepsilon_{k}^{\text{c}} \mid \Delta \mathbf{x}_{k}, \mathcal{M}_{k})$ is the likelihood distribution of the $\varepsilon_{k}^{\text{c}}$. This distribution is hard to determine in practice because 
(i) it is affected by many factors which are hard to model, (ii) the detection and matching of features as well as the motion estimation are actually algorithms that has no closed form expressions which makes the propagation of these errors to the motion increments infeasible. 

Consequently, we propose the use of a feedforward Neural Network (NN) to estimate the error $\varepsilon_{k}^{\text{c}}$ using the output of a monocular VO algorithm $\Delta \mathbf{x}_{k}$ as well as the matched features pairs $\mathcal{M}_{k}$.   

%%%%%%%%%%%%%%%%%%%%%%%%%%%%%%%%%%%%%%%%%%%%%%%%%%%%%%%%%%%%%%%%%%%%%%%%%
\section{Drift Reduction using Neural Networks}
\label{sec:data_driven}
In this section, the NN used to estimate the noise component $\varepsilon^{\text{c}_{k}}$ is described. A monocular VO algorithm estimates the incremental motion of the vehicle up to a scale which can be computed using an external sensor such as a wheel encoder or an IMU. the proposed work is only concerned with estimating the error component for the orientation increment $\Delta \theta$. The overall pipeline of the VO algorithm with the proposed Drift Reducing Neural Network (DRNN) is shown in Fig.~\ref{fig:overall_pipeline}.

The objective of the proposed DRNN is to estimate a refined incremental orientation $\Delta \theta$ using the output of the monocular VO as well as information about the matched features $\mathcal{M}_{k}$. As disucssed in the previous section, the error in the matched features result in errors in the estimated motion increments by the VO. Such errors can be detected using the motion of the features in a given frame $I(\mathbf{x}_{k})$ with respect to $I(\mathbf{x}_{k-1})$
\begin{equation}
	\underbrace{\begin{bmatrix}
		\Delta u^{i}_{k} \\ \Delta v^{i}_{k}
	\end{bmatrix}}_{\Delta f^{i}_{k}} = f_{k}^{i} - f_{k-1}^{i},
\end{equation}
where $\Delta u$ and $\Delta v$ are the change in the position of the feature in the successive frames. Therefore, the DRNN can be trained using the position change of the features. 

The space of the matched features in the successive frames is large and many different types of error can occur consequently, training the DRNN using all the matched features would require a very large amount of training samples which would be infeasible during the training phase of the DRNN. Alternatively, we use some of the statistical moments of the set $\{\Delta f_{k}^{i}\}_{i=1}^{N_f}$ to train the DRNN to reduce the error. In this paper, the information about the change in the positions is captured using three statistical moments (mean $\mu_{k}$, variance $\nu_{k}$, skewness $\kappa_{k}$) in addition to the Root Mean Square (RMS) $\rho_{k}$.

As for the orientation increment, the orientation representation used is the unit quaternions. It is well-known that the rotation space $\mathbb{M}$ is not an Euclidean space. 
%($\mathbb{M} \ncong \mathbb{R}^{3}$). 
However, the tangent space of the rotation space is an Euclidean space ($\mathbb{R}^{3}$). Consequently, to avoid discontinuities in the cost function of the DRNN, the orientation deviation in the tangent space $\xi := \exp_{\text{q}}(\Delta \theta) \in \mathbb{R}^{3}$ is used where $\exp_{\text{q}}: \mathbb{M} \rightarrow \mathbb{R}^{3}$ is the exponential map that maps from the quaternion space to the tangent space.

Finally, the input layer of the DRNN is defined as $[(\xi^{-}_{k})^{\top}, \mu_{k}, \nu_{k} , \kappa_{k}, \rho_{k}]^{\top}$, where $\xi^{-}_{k} := \exp_{\text{q}}(\Delta \theta^{-}_{k})$ is the orientation deviation of the orientation increment estimated by the VO algorithm. The output layer of the DRNN is the refined orientation increment $\xi^{+}_{k}$. The overall position and orientation of the vehicle can finally be computed as 
\begin{align}
    \theta^{+}_{k} &= \theta^{+}_{k-1} \odot \Delta \theta^{+}_{k} \\
    \mathbf{p}_{k} &= \mathbf{p}_{k-1} + R(\theta^{+}_{k}) \Delta \mathbf{p}_{k}
\end{align}
where $\odot$ denotes the quaternion product, and $R: \mathbb{M} \rightarrow SO(3)$ is a mapping from quaternion group to the special orthogonal group. Furthermore, to make sure that the DRNN does not overfit to the training data and is actually capable of reducing the VO errors during operation, a Bayesian regularization approach is used ~\cite{Foresee1997GaussNewtonAT}.

%%%%%%%%%%%%%%%%%%%%%%%%%%%%%%%%%%%%%%%%%%%%%%%%%%%%%%%%%%%%%%%%%%%%%%%%%

\section{Experimental Work}
In this section we discuss the details of the implemented monocular VO odometry algorithm and the executed experimental work in order to validate our approach. Using Mathworks' Matlab, a feature-based monocular VO algorithm is developed  using Speeded Up Robust Features (SURF) detector~\cite{article},~\cite{surf}. The features are then matched using the approach introduced in~\cite{features2},
where only unique features are matched by using forward-backward matching. 

The monocular VO implementation estimates the motion $\Delta \mathbf{x}$ using the epipolar constraint  between frames 
\begin{equation}
	\begin{bmatrix}
		x^{i}_{k-1} & y^{i}_{k-1}
	\end{bmatrix} \mathbf{E} \begin{bmatrix}
		x^{i}_{k} \\ y^{i}_{k} \\
	\end{bmatrix} = 0, \ \ \forall \ 1 \leq i \leq N_f, 
\end{equation}
where $(x^{i}, y^{i})$ is the position of the $i$-th feature in the camera frame, and $\mathbf{E} \in \mathbb{R}^{3 \times 3}$ is the essential matrix for the calibrated camera~\cite{hartley2003multiple}.
The essential matrix $\mathbf{E}$ can be estimated using the five-point algorithm~\cite{features3}. Furthermore, an outliers rejection approach was applied based on the M-estimator sample consensus (MSAC) algorithm~\cite{features4}.
The essential matrix $\mathbf{E}$ can then be decomposed to the translation (up to a scale) and rotation of the camera between two successive frames~\cite{hartley2003multiple}. 

Since the monocular VO algorithm estimates the motion up to a scale, here we compute this scale using the ground-truth data for both the VO with and without the DRNN as we are only concerned with reducing the orientation drift in the VO output.
This does not affect the validation of the proposed approach since the scale is computed using the same method for both outputs.

The outcome of the VO is passed to a feed-forward NN with 3 layers. The input layer consists of 11 neurons for the inputs discussed in the previous section. The hidden and output layers consist of 30 and 3 neurons respectively.
The activation function used is a sigmoid function and
the weights of the NN are optimized using the Levenberg-Marquardt optimization algorithm~\cite{trainNN} and the training is done using 1500 training epochs.
%a minimum performance gradient threshold of $5^{-7}$, and Marquardt adjustment parameter of $5^{-3}$. Finally the training was made using 
%1500 training epochs.

Using the KITTI data set, the NN is trained with the first $60\%$ percent of sequences 1, 2, 5, 6, 7, 8, and 9 and is tested on the remaining $40\%$ percent of each sequence. Sequence $4$ is used completely for training as it is too short for partitioning. Finally, sequence $10$ is completely used for testing the DRNN. The training is accomplished by using the ground-truth data as targets for the NN.

The performance of the DRNN is validated and its output is compared to that of the monocular VO algorithm without the DRNN. For both outputs, the Root Mean Squared Error (RMSE) in translation and orientation are calculated using the ground-truth data provided for each sequence. The orientation error is computed as
\begin{equation}
R^{\text{e}, \pm}_k = R(\theta^{\text{GT}}_k)^{-1} R(\theta^{\pm}_k) 
\end {equation}
where $ R^{\text{e, +}}_k, R^{\text{e, -}}_k \in SO(3)$ represents the error rotation matrix for the VO with and without the proposed DRNN, $\theta^{\text{GT}}_k$ represents the ground truth orientation in as a quaternion.

\label{sec:experimental_work}

%%%%%%%%%%%%%%%%%%%%%%%%%%%%%%%%%%%%%%%%%%%%%%%%%%%%%%%%%%%%%%%%%%%%%%%%%

\section{Results and Discussion}
 \label{sec:results_discussion}

 \begin{figure}[t]
 	\centering
	\includegraphics[trim={0cm 0cm 0cm 0.85cm}, clip, width=0.95\linewidth]{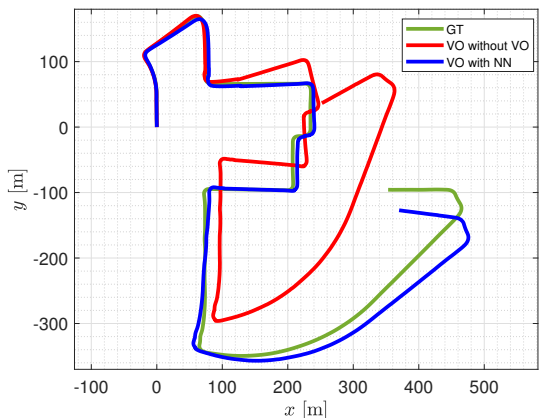}  
	\caption{The computed trajectory by the VO with and without the DRNN along with the ground-truth trajectory of the testing portion of sequence 0 of KITTI dataset.}
	\label{fig:seq1}
\end{figure}
\begin{figure}[t]
	\centering
	\includegraphics[width=0.95\linewidth]{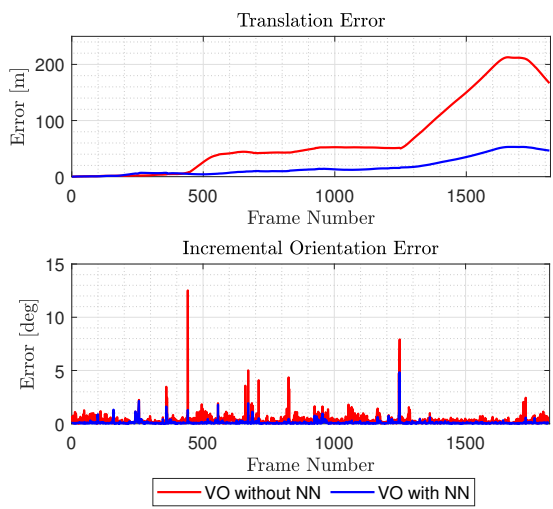}  
	\caption{The translation error in the position estimation of the vehicle in sequence 0 (top) and the incremental orientation error at every frame (bottom).}
	\label{fig:seq1error}
\end{figure}

In this section the results of the testing scenarios are  discussed as mentioned in Section \ref{sec:experimental_work}. Fig.~\ref{fig:seq1} shows the estimated trajectory by the VO algorithm with and without the proposed DRNN for the testing part of sequence 0 as well as the ground-truth trajectory. As can be seen in the figure, the computed trajectory using the VO algorithm without the DRNN suffers from significantly more drift compared to the VO algorithm with the DRNN. 

 \begin{figure}[t]
 	\centering
	\includegraphics[trim={0cm 0cm 0cm 0.85cm}, clip, width=0.95\linewidth]{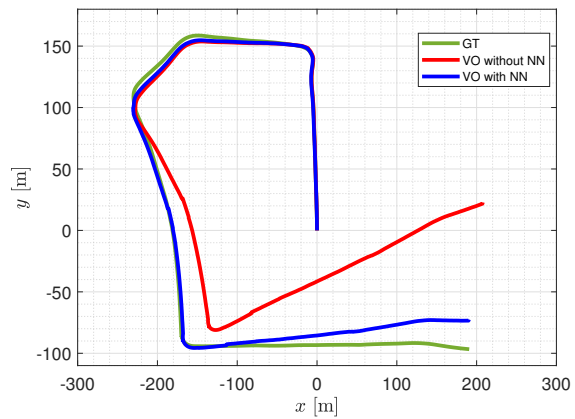}  
	\caption{The computed trajectory by the VO with and without the DRNN along with the ground-truth trajectory of the testing portion of sequence 5 of KITTI dataset.}
	\label{fig:seq5}
\end{figure}

\begin{figure}[t]
	\centering
	\includegraphics[trim={0cm 0.2cm 0cm 0cm}, clip, width=0.95\linewidth]{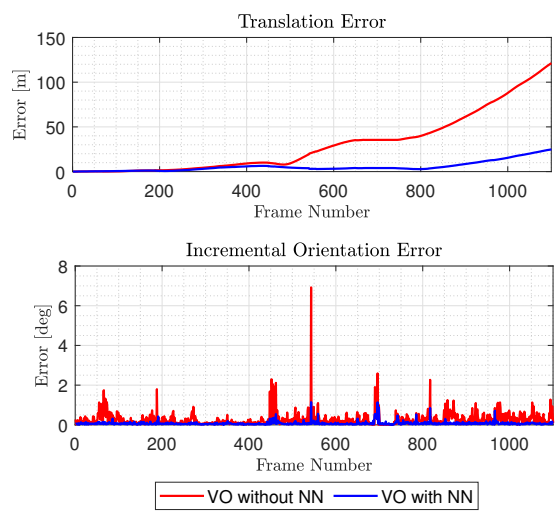}  
	\caption{The translation error in the position estimation of the vehicle in sequence 5 (top) and the incremental orientation error at every frame (bottom).}
	\label{fig:seq5error}
\end{figure}

Fig.~\ref{fig:seq1error} shows the translation error as well as the incremental orientation error for sequence 0. Using the proposed DRNN, the estimation of the incremental orientation change was persistently more accurate over the sequence frames which led to a better overall pose estimation. Although the VO output was relatively accurate at the beginning of the path however, with every turn that the vehicle took, the amount of drift in the estimated orientation increased which led to a root mean translation error of $94.19$ m compared to only $23.35$ m when using the DRNN which amounts to a $75\%$ improvement. This can be attributed to the enhanced orientation estimation which was enhanced from a RMSE of $15.22$ degrees to only $5.03$ degrees ($66\%$) when using the DRNN.

Fig.~\ref{fig:seq5} shows the results for the testing part of sequence 5. This sequence did not contain as many turns as sequence 0, which resulted in a slower drift compared to sequence 0 and a less error in the overall sequence (shown in Fig.~\ref{fig:seq5error}). That being said, by the end of the trajectory the VO output still suffered from a larger amount of drift compared to the VO output with the DRNN. In the case of using the DRNN, the RMSE in orientation for sequence 5 was reduced from $12.44$ degrees to $3.2$ degrees ($74\%$ improvement) which resulted in $82\%$ enhancement in the position accuracy. Here, the DRNN output was also persistently better than that of the VO output. Furthermore, notice that the DRNN managed to reduce the incremental orientation error at the error spikes. These spikes can lead to a significant increase in the overall error in the path due to the reliance on integration. 

 \begin{figure}[t]
 	\centering
	\includegraphics[trim={0cm 0cm 0cm 0.7cm}, clip, width=0.95\linewidth]{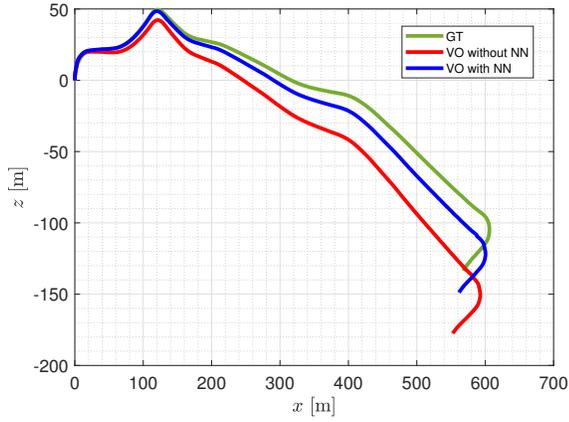}  
	\caption{The computed trajectory by the VO with and without the DRNN along with the ground-truth trajectory of the testing portion of sequence 9 of KITTI dataset.}
	\label{fig:seq9}
\end{figure}

\begin{figure}[t]
	\centering
	\includegraphics[width=0.95\linewidth]{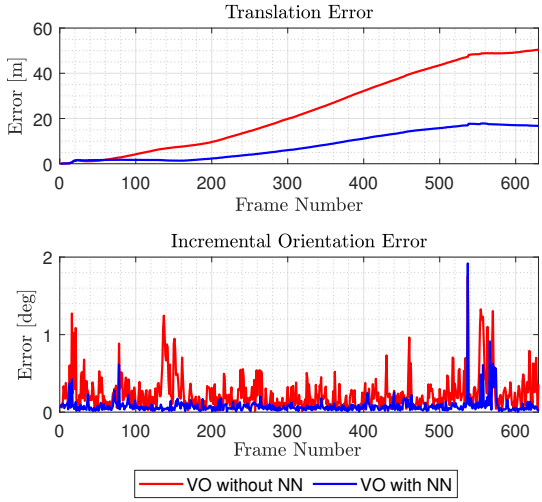}  
	\caption{The translation error in the position estimation of the vehicle in sequence 9 (top) and the incremental orientation error at every frame (bottom).}
	\label{fig:seq9error}
\end{figure}

Similarily, Fig.~\ref{fig:seq9} and~\ref{fig:seq9error} show the results for sequence 9 testing part of KITTI dataset which are similar to those of sequence 5. In this case, the orientation RMSE was reduced from $7.11$ degress to $2.08$ degrees ($70\%$ improvement) and the translation RMSE was reduced from $29.66$ m to $10.34$ m ($65\%$ improvement). 

\begin{figure}[t]
	\centering
	\includegraphics[trim={0cm 0cm 0cm 0.7cm}, clip, width=0.95\linewidth]{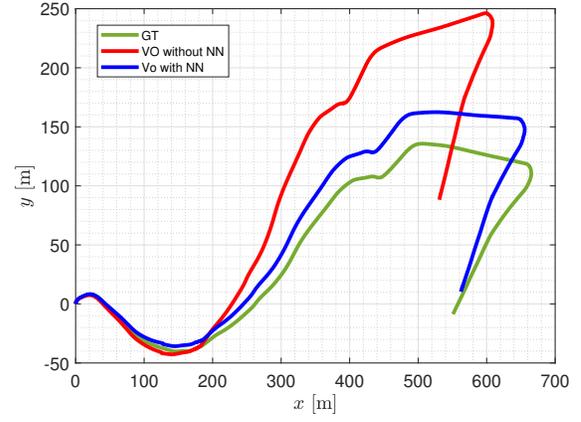}  
	\caption{The computed trajectory by the VO with and without the DRNN along with the ground-truth trajectory of the testing portion of sequence 10 of KITTI dataset.}
	\label{fig:seq10}
\end{figure}

\begin{figure}[t]
	\centering
	\includegraphics[width=0.95\linewidth]{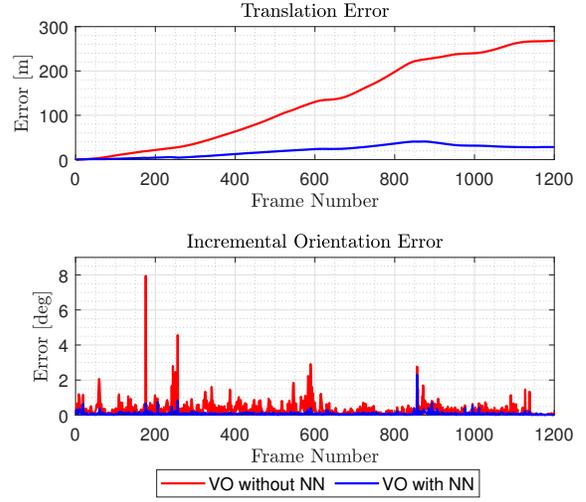}  
	\caption{The translation error in the position estimation of the vehicle in sequence 10 (top) and the incremental orientation error at every frame (bottom).}
	\label{fig:seq10error}
\end{figure} 

Fig.~\ref{fig:seq10} and~\ref{fig:seq10error} show the estimation results of the complete trajectory of sequence 10 as well as the translation and incremental orientation errors. Even though the sequence was completely used for testing without partitioning, the use of the DRNN resulted in significant enhancement in the translation and orientation estimation accuracy. In addition to the previous tests using the partitioned sequences, this sequences show the practical feasibility of the proposed DRNN. The idea behind the DRNN is that the drift can be reduced through training the NN during the development phase of the vehicle and then using the DRNN during operation to reduce the drift in the VO estimate. For sequence $10$, the orientation RMSE was reduced from $29.06$ degrees to $7.02$ degrees ($76\%$ improvement) which then led to a reduction in translation error from $160.56$ m to $23.85$ m ($85\%$ improvement).

Finally, Table~\ref{tab:results} shows the overall results for the used test sequences. The use of the DRNN resulted in a significant reduction in the translation and orientation errors for all of the testing sequences which shows that through using the proposed approach, the performance of the VO algorithm can be significantly enhanced.   
 
 \begin{table}[ht]
\centering
\caption{RMSE Values for the Kitti Dataset Sequences}
\label{tab:results}
\renewcommand{\arraystretch}{1.1}
\begin{tabular}{cccccc}
\hline \hline
\multirow{2}{*}{Seq. No.} & \multicolumn{2}{c}{Rotation {[}deg{]}} & \multicolumn{2}{c}{Translation {[}m{]}} & \multirow{2}{*}{Distance [m]} \\ \cline{2-5}
                      & NN                     & VO            & NN                     & VO             &                           \\ \hline
0                     & \textbf{5.03}          & 15.22         & \textbf{23.35}         & 94.19          & 1648                      \\ \hline
2                     & \textbf{8.34}          & 23.73         & \textbf{39.65}         & 93.46          & 2108                      \\ \hline
5                     & \textbf{3.2}           & 12.44         & \textbf{8.05}          & 45.36          & 958                       \\ \hline
6                     & \textbf{3.79}          & 15.26         & \textbf{6.96}          & 49.88          & 491                       \\ \hline
7                     & \textbf{1.83}          & 3.77          & \textbf{4.08}          & 5.44           & 238                       \\ \hline
8                     & \textbf{6.88}          & 16.8          & \textbf{35.86}         & 81.72          & 1348                      \\ \hline
9                     & \textbf{2.08}          & 7.11          & \textbf{10.34}         & 29.66          & 706                       \\ \hline
10                    & \textbf{7.02}          & 29.06         & \textbf{23.85}         & 160.56         & 919                       \\ \hline \hline
\end{tabular}
\end{table}

%%%%%%%%%%%%%%%%%%%%%%%%%%%%%%%%%%%%%%%%%%%%%%%%%%%%%%%%%%%%%%%%%%%%%%%%%
\section{Conclusion and future work}
\label{sec:conclusion}
In this paper, a drift reduction approach for the orientation estimation in VO using a neural network is proposed. The proposed approach estimates the error which is correlated to the errors in the feature detection and matching algorithms and consequently is able to refine the incremental orientation estimation of the VO odometry. Through training the neural network during the development of the vehicle, the performance of the VO can be significantly enhanced using the proposed approach. our proposed DRNN was validated using KITTI dataset, where 7 sequences  where partitioned to training and testing sequences for validation. Furthermore, a complete sequence was used for further validation of the algorithm. The results show the efficacy of the proposed DRNN in significantly reducing the drift in the VO output.

In the future work, the proposed approach can be extended to work with different types of cameras such as RGB-D or stereo cameras as well as different VO algorithms such as direct VO. Furthermore, the effect of the proposed approach can also be integrated with visual simultaneous localization and mapping algorithms to achieve more accurate results and avoid significant drift in the pose estimates. 
%%%%%%%%%%%%%%%%%%%%%%%%%%%%%%%%%%%%%%%%%%%%%%%%%%%%%%%%%%%%%%%%%%%%%%%%%

\bibliographystyle{IEEEtran}
\bibliography{journal}

\end{document}